\documentclass[times,twocolumn,final]{elsarticle}

\usepackage{caption}

\usepackage{xcolor}
\usepackage{lettrine}
\usepackage{natbib}

\usepackage{graphicx}
\usepackage{epstopdf}
\usepackage{generic}
\usepackage{amsmath,amssymb,amsfonts}
\usepackage{algorithmic}
\usepackage{graphicx}
\usepackage{algorithm,algorithmic}
\usepackage{hyperref}
\hypersetup{hidelinks}
\usepackage{textcomp}
\usepackage{subcaption}

\usepackage{stfloats}
\usepackage{makecell}
\usepackage{multirow}
\usepackage{threeparttable}
\usepackage[T1]{fontenc}

\begin{document}
\bibliographystyle{IEEEtran}

\title{The Age-specific Alzheimer \textquotesingle s  Disease Prediction with Characteristic Constraints in Nonuniform Time Span
}
\author[Hqu]{Xin Hong}
\ead{email: xinhong@hqu.edu.cn}

\author[Hqu]{Kaifeng Huang}

\address[Hqu]{Department of Software, School of Computer Science and Technology, Huaqiao University, China}

\begin{abstract}
Alzheimer's disease is a debilitating disorder marked by a decline in cognitive function. Timely identification of the disease is essential for the development of personalized treatment strategies that aim to mitigate its progression. 
The application of generated images for the prediction of Alzheimer's disease poses challenges, particularly in accurately representing the disease's characteristics when input sequences are captured at irregular time intervals. This study presents an innovative methodology for sequential image generation, guided by quantitative metrics, to maintain the essential features indicative of disease progression. Furthermore, an age-scaling factor is integrated into the process to produce age-specific MRI images, facilitating the prediction of advanced stages of the disease. The results obtained from the ablation study suggest that the inclusion of quantitative metrics significantly improves the accuracy of MRI image synthesis. Furthermore, the application of age-scaled pixel loss contributed to the enhanced iterative generation of MRI images. In terms of long-term disease prognosis, the Structural Similarity Index reached a peak value of 0.882, indicating a substantial degree of similarity in the synthesized images.

\end{abstract}

\begin{keyword}
Alzheimer's Disease, GAN, MRI, Temporal, Quantitative Indicator
\end{keyword}

\maketitle

\section{Introduction}
\label{sec:introduction}
Alzheimer’s disease (AD) is a neurodegenerative disorder that imposes substantial cognitive and financial burdens on individuals and their families \cite{cao2018brits}. The prediction of the disease and the implementation of effective therapeutic strategies are crucial in mitigating the advancement of AD \cite{andrade2009prevention}.Utilizing temporal imaging for early prediction is imperative for comprehending the disease's progression and assessing treatment efficacy \cite{wang2020clinical,zhu2021long}.

In the domain of brain image prediction, conventional non-deep learning approaches \cite{burgos2014attenuation,papadimitroulas2013investigation} demonstrate limited effectiveness \cite{pan2021disease}. Over the past few years, deep learning techniques have achieved significant advancements in the  field of computer vision and medical imaging prediction \cite{goodfellow2020generative,ramesh2022hierarchical}.Prediction methods for imaging can be broadly classified into three categories: cross-modal, Vector-to-Image, and temporal image prediction. 

Cross-modal techniques forecast cross-modal imaging data by establishing image associations across different modalities \cite{kong2021breaking,li2014deep,nie2017medical,pan2021collaborative}.While these methods facilitate lateral mapping between cross-modal images at the same time scale, they are less effective in predicting future temporal images. 

Vector-to-Image methods involve mapping random vectors to image space\cite{kwon2019generation,pinaya2022brain,ramesh2022hierarchical}.Despite their ability to generate images, these methods do not guaranty that the predicted temporal images originate from the same sample. 

Temporal methods play a crucial role in predicting future image sequences by allowing the model to capture patterns in temporal changes within images\cite{elazab2020gp,fan2022tr,ravi2019degenerative}. The endeavor of temporal image prediction encounters two primary challenges. Firstly, within datasets comprising MRI image sequences, there is a greater prevalence of short-term sequences compared to long-term sequences. This disparity adversely impacts the accuracy of predictions pertaining to long-term sequences. 
Secondly, it is crucial to maintain disease-related characteristics in the prediction of MRI images\cite{rosenberg1984rey,kanekiyo2014apoe,tombaugh1992mini,chetelat2003mild,liu2016relationship, shaw2009cerebrospinal,rahimi2014prevalence}. Several current methodologies\cite{xia2019consistent,xia2021learning, yoon2023sadm}  utilize disease states and clinical information as prior knowledge to improve the model's comprehension of the disease.

Nonetheless, challenges arise from the irregular sequences of various input time intervals, such as specific assessments or imaging modalities, which hinder the effective application of temporal predictions. To address these challenges, this research introduces the Image Temporal Generative Adversarial Network (T-GAN), which comprises a cross-attention generator, an adversarial discriminator, and a quantitative indicator discriminator. The primary objective of this study is to facilitate the early detection of Alzheimer's disease, with the key contributions outlined as follows:
\begin{itemize}

\item The generator's ability to discern temporal dynamics in image features is enhanced through the integration of cross-attention mechanisms that incorporate age-related constraints.

\item Additionally, the pixel loss function has been adjusted by incorporating an age-scaling factor. This modification serves to improve the model's emphasis on generating images that originate from non-uniform temporal sequences;
 
\item Moreover, the preservation of disease characteristics within the sequence of generated images is achieved through the implementation of a quantitative indicator discriminator;

\item By employing existing quantitative indicators to train the adversarial discriminator for imaging purposes, the model demonstrates the capability to predict future MRI images, even in the absence of quantitative indicator data as input.
\end{itemize} 

\section{RELATED WORK}

The early prediction of Alzheimer's disease (AD) can be facilitated through the analysis of brain imaging. Sequential medical imaging data consist of a collection of medical images obtained from patients at various time intervals, which are essential for investigating the progression of the disease over time.

\subsection{Sequential Brain Image Prediction}
A multitude of factors influences the nonuniform temporal scaling of sequences within brain imaging data, potentially leading to biases in modeling and subsequently impacting the predictive accuracy of brain images.

A wide range of sampling-based methodologies has been extensively utilized to address the challenges associated with imbalanced time interval sequential data \cite{cao2013integrated,liang2013effective,thai2010cost}. For example, the hybrid sampling strategy \cite{liang2013effective} effectively mitigates computational expenses and training duration linked to oversampling by leveraging a limited number of positive samples while simultaneously improving the undersampling process through the integration of a larger volume of negative samples. Integrated Oversampling (INOS) \cite{cao2013integrated} merges Enhanced Structured Preserving Oversampling (ESPO) with interpolation-based oversampling techniques. Furthermore, recurrent neural networks, in conjunction with oversampling methods, have been utilized to generate data from limited datasets \cite{gong2016model}. In contrast, the integration of resampling techniques with cost-sensitive learning has also been investigated \cite{thai2010cost}. Nonetheless, it is essential to recognize that sampling-based approaches may lead to the loss of critical data or exacerbate the issue of model overfitting \cite{liang2013effective}.

In the context of temporal brain image prediction tasks, a combination of Variational Autoencoders (VAE) and Wasserstein Generative Adversarial Networks (GAN) was employed to forecast temporal MRI images by utilizing random noise that conforms to a Gaussian distribution \cite{kwon2019generation}. However, this approach was unable to ascertain the age corresponding to the generated prediction images. Conversely, TR-GAN \cite{fan2022tr} implemented an iterative network with a six-month interval for MRI image prediction. While this method demonstrated improved performance in short-term image predictions, it posed a risk of information loss in long-term predictions due to the potential for excessive iterations. Similarly, the Sequence-Aware Diffusion Model (SADM) \cite{yoon2023sadm} was developed to investigate the temporal dynamics and dependencies of sequential processes in diffusion models for medical image generation. This research addresses the limitations associated with the disease features in the predicted images. Furthermore, the LD-GAN \cite{ning2020ldgan} not only focuses on predicting MRI images by establishing a bidirectional mapping between two adjacent time-point MRI images but also emphasizes the prediction of clinical scores. Although this method was primarily designed for interpolation between adjacent images, it was limited in its capacity for long-term image prediction.

In tackling the complexities associated with varying temporal spans within datasets, T-GAN employs a mechanism that integrates age and image features, thereby enhancing the generator's capacity to understand temporal dynamics.

\subsection{Preserving Disease Features in Brain Image Prediction}

Research on controlled image prediction \cite{huang2020pfa,mirza2014conditional} has highlighted the importance of maintaining the disease characteristics of patients when generating temporal brain images for predictive applications. 

Quantitative indicators of disease and clinical information have been utilized as diagnostic criteria for Alzheimer's disease (AD) \cite{rahimi2014prevalence}. Their integration into temporal image prediction tasks has effectively enhanced the model's capacity to forecast disease-related features. The Conditional Generative Adversarial Network (GAN) \cite{mirza2014conditional} is a well-established, constraint-based generative framework that has been extensively employed to predict controlled brain images. Notably, the Age-ACGAN \cite{kan2020age} was designed to generate computed tomography (CT) brain images of children for specific developmental stages; however, it did not adequately maintain the integrity of disease features during image prediction. Furthermore, the Enhanced Conditional GAN \cite{xia2019consistent,xia2021learning} managed to regulate the disease characteristics of predicted magnetic resonance imaging (MRI) images by utilizing health state and age as constraints. However, these methods focus solely on the disease state when generating images, overlooking the pathological constraints of other diagnostic indicators related to the condition. Furthermore, diffusion models \cite{dhariwal2021diffusion} have been employed in the realm of brain imaging prediction \cite{pinaya2022brain}, incorporating covariates to account for factors such as age, gender, and  brain structure volumes. It is essential to recognize that these models require the incorporation of prior information. The characteristics of diseases are often not readily available in large-scale, real-world screening programs, which complicates the accuracy of predictions.

We compared five GAN models for temporal AD brain image prediction. The Age-ACGAN [36] generated age-conditioned high-resolution CT images. The CAAE [52] was adapted to predict facial aging by incorporating gender and age as input conditions. The Pix2Pix model handled two formats: Pix2Pix-r combined the original input with a random variable, while Pix2Pix-A [40] used the target image’s age instead. The TR-GAN [17], the latest model, generated images iteratively every six months in this experiment.

In response to the identified challenges, T-GAN employs enhanced supervision of quantitative metrics in conjunction with image data during the training phase. This approach allows T-GAN to predict images exhibiting disease-related characteristics without requiring patient metrics as input.

\section{METHOD}

\begin{figure*}[!t]
\centerline{\includegraphics[width=1\linewidth]{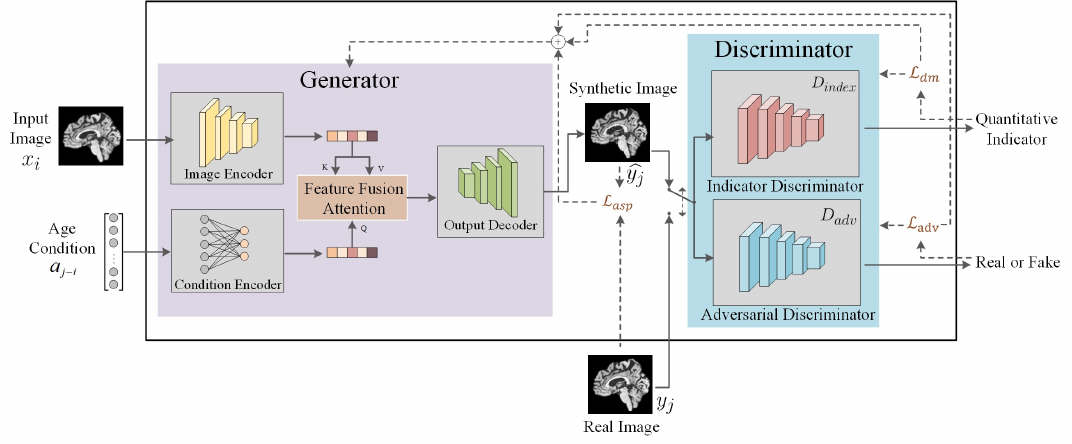}}
\caption{Simplified diagram of the model's structure. (a) Generator: The generator encodes and fuses the input images and ages features, decoding them into synthetic images. (b) Discriminator: The generated and real images are fed into the discriminator to obtain the quantitative indicators and image discriminator.}
\label{fig_network}
\end{figure*}

In order to improve the efficiency of image prediction, this study introduces a new temporal image generative adversarial network (T-GAN), which comprises a generator and two discriminators, as depicted in Fig. \ref{fig_network}. The network concurrently tackles the prediction of images and quantitative indicators by integrating an additional discriminator specifically for quantitative indicators, which serves to regulate the image generator. The T-GAN incorporates a cross-attention module designed to integrate sequential image features while accounting for age-related constraints, and it effectively captures sequential disease characteristics through the use of quantitative indicators. Furthermore, the pixel loss is modified by an age-scaling factor to enhance the model's focus on the irregular time intervals present in the input sequences. 
This architectural choice guaranties that the generated images retain the characteristics of the disease, thereby enhancing the accuracy of the predictions.

\subsection{Formulation of T-GAN}

\begin{figure}[!t] 
\centerline{\includegraphics[width=1\linewidth]{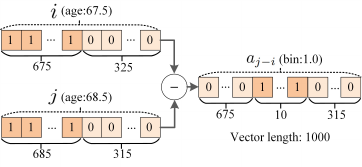}}
 \caption{Age Encoding Method. The age feature is encoded as a binary vector of length 1000, with a maximum age of 100.0 years, as the maximum age of samples in the dataset is 97.3. In this case, the age of the $i$th-period for the sample is 67.5 years, where the first 675 elements of the encoding vector are set to 1, and the remaining 325 elements are set to 0. Similarly, the encoding method can be applied to the $j$th-period. Subtracting the encoding vectors of two periods obtains the age difference vector$x_a$.}
\label{fig_age_code}
\end{figure}

T-GAN seeks to establish a relationship within a Generative Adversarial Network (GAN) framework between quantitative indicators and images while ensuring that disease-related characteristics are maintained in the generated images. Let $\mathcal{I}$ denote the distribution of images within the dataset, $\mathcal{A}$ represent the distribution of ages, and $\mathcal{M}$ signify the distribution of quantitative indicators associated with the images. For a given sample observed at two distinct time points, the corresponding pair of images is represented as $({x_i, y_j}) \in \mathcal{I}$, where $({i, j}) \in \mathcal{A}$ and $i < j$. The image $\widehat{y_j}$ generated by the model can be expressed as $\widehat{y_j} = G(x_i, a_{j-i})$, with $a_{j-i}$ encoding the temporal difference between $i$ and $j$. The method of encoding is elaborated in Fig. \ref{fig_age_code}. The objective of the generative model is to produce images that closely approximate the target images. The output of the adversarial discriminator, denoted as $D_{adv}({y_j}|G(x_i, a_{j-i}))$, can be formulated in Eq.(\eqref{eq_e}):

\begin{equation}
    \begin{split}
        \widehat{\mathbb{E}}=\operatorname*{min}_{\mathbb{E}}|D_{adv}(G(x_i,a_{j-i}))-D_{adv}({y_j})|
    \end{split}
   \label{eq_e}
\end{equation}

In conjunction with the aforementioned discriminator for generated images, T-GAN integrates an indicator discriminator branch, referred to as $D_{indicator}$, which guaranties the alignment between the quantitative indicators of $\widehat{y_j}$ and $y_j$. The role of $D_{indicator}$ is to preserve sample-specific disease-related characteristics within $\widehat{y_j}$. The output produced by the indicator discriminator is denoted as $D_{indicator}({y_j}|G(x_i, a_{j-i}))$. The revised expectation can be articulated in Eq.(\ref{eq_de}):

\begin{equation}
    \label{eq_de}
    \begin{split}
        \widehat{\mathbb{E}} = \min_{\mathbb{E}} \left( |D_{adv}(G(x_i,a_{j-i}))-D_{adv}({y_j})|+ \right. \\
        \left. |D_{indicator}(G(x_i, a_{j-i})) - D_{indicator}(y_j)| \right)
    \end{split}
\end{equation}

Equation (\ref{eq_de}) delineates the fundamental aims of T-GAN, which include: (a) the development of a robust temporal image generator that utilizes age-related conditions for the purpose of image prediction, (b) the establishment of a generative adversarial network that incorporates discriminators for assessing image authenticity, and (c) the improvement of disease feature retention in the generated images by implementing a diagnostic indicator discriminator for supervisory guidance.

\begin{figure*}[htbp]    \centerline{\includegraphics[width=18cm,height=8cm]{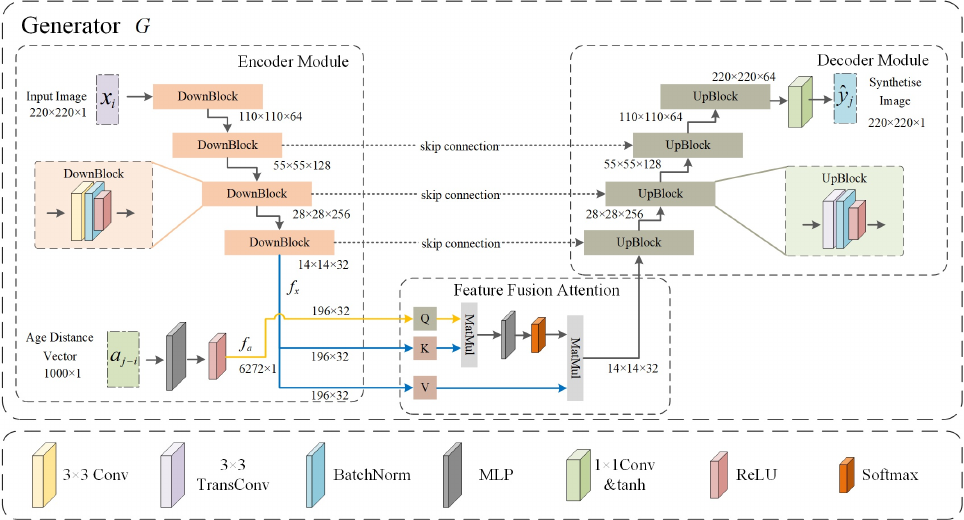}}
 \caption{The generator architecture diagram illustrates the utilization of the U-Net as the foundational network to construct the generator. (a) Image Encoder: Four downsampling convolutional layers extract pertinent features from the input image. (b) Condition Encoder: A multilayer perceptron is a conditional encoder for age-conditional information. (c) Feature Fusion Attention: The amalgamation of features is executed by the attention mechanism during the feature fusion process.(d) Output Decoder: The decoder module comprises four upsampling deconvolutional layers responsible for the final image's reconstruction. Ultimately, the resultant image is acquired by subjecting the output through a convolutional layer, succeeded by a tanh activation function.
    }
    \label{fig_gen}
\end{figure*}

\subsection{Generator}

The Generator model enhances brain imaging sequence generation by incorporating age-related constraints. It uses a brain image \( x_i \) and an age conditional vector \( x_a \) as inputs. The T-GAN architecture employs an attention mechanism to merge these vectors, improving the generator's sensitivity to age-related features. The decoder then produces the target image. The generator comprises three main components: an encoder module, a feature fusion module, and a decoder module, as illustrated in Fig. \ref{fig_gen}.

\subsubsection{Encoder Module}

The encoder module includes two components: the image encoder and the conditional encoder.The image encoder extracts features from the brain image slice \( x_i \) and encodes them into a low-dimensional latent matrix. Meanwhile, the conditional encoder processes the age vector \( x_a \), aligning it with the extracted image features.

\subsubsection{Feature Fusion Attention}

When integrating image features with age-conditional features, dimensional discrepancies can arise, as image-derived features often outnumber age-conditional features. T-GAN addresses this by using an encoder to enhance the dimensionality of age-conditional features, which are then combined with image features through an attention mechanism\cite{vaswani2017attention}. In this procedure, T-GAN flattens the features from each channel of the image features, designating them as Key (K) and Value (V). The features from the age-conditional constraint are utilized as Query (Q). T-GAN derives the final output by assessing the correlation between the Query and Key, extracting the corresponding weights, and applying these weights to the Value, thereby enhancing the model's understanding of age-conditional features. Let \( f_a \) represent the feature vector corresponding to age condition encoding, \( f_x \) denote the feature vector of the input image, and \( W_q, W_k, W_v \) signify the respective weight matrices for Query, Key, and Value.The process is delineated in Eq.(\ref{eq_qkv}):

\begin{equation}
    \label{eq_qkv}
    \begin{aligned}Q&=f_aW_q\\K&=f_xW_k\\V&=f_xW_v\end{aligned}
\end{equation}

Subsequently, we calculate the correlation between the Query and Key, from which we derive the weight coefficients using the Softmax function. These coefficients are then applied to the Value to produce the final output. The procedure is outlined in Eq.(\ref{eq_qkv2}):
\begin{equation}
   \label{eq_qkv2}
\operatorname{Attention}(Q,K,V)=\operatorname{Softmax}\Bigg(\frac{Q\cdot K^T}{\sqrt{d_k}}\Bigg)\cdotp V
\end{equation}
where $d_k$ is the dimension of $K$ and $K^T$ is the transpose matrix of $K$.

\subsubsection{Decoder Module}

At this stage, the decoder module leverages fused features and executes upsampling to produce the final output image. To improve the quality of the generated images, T-GAN incorporates skip connections \cite{ronneberger2015u} that link lower-level features from the encoder with higher-level features from the decoder.

\subsection{Discriminator}
\begin{figure*}[htbp]
    \centerline{\includegraphics[width=18cm,height=9cm]{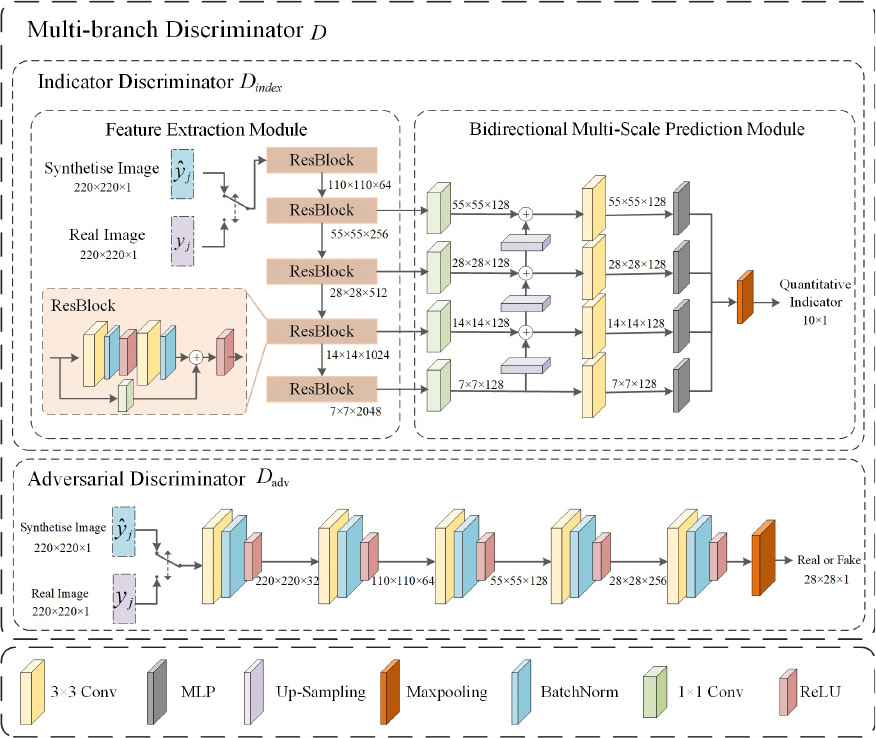}}
    \caption{Simplified diagram of the model's discriminator. 
    (a) Indicator Discriminator: The adversarial discriminator consists of five convolutional layers and outputs the authenticity of each pixel value in the image. 
    (b) Adversarial Discriminator: The Adversarial Discriminator using the Feature Pyramid Network produces precise quantitative indicators by calculating the average of outputs from each scale in the network.}
    \label{fig_disc}
\end{figure*}

The discriminator model introduces a novel image generation method guided by the constraints of the indicator discriminator. It uses metric loss supervision to evaluate disease-relevant feature retention. As shown in Fig. \ref{fig_disc}, it consists of the adversarial discriminator module $D_{adv}$ and the indicator discriminator module $D_{indicator}$.

\subsubsection{Indicator Discriminator}
The discriminator uses a network to predict clinical quantitative indicators from input images. It includes a feature extraction module and a bidirectional multi-scale prediction module that processes multi-scale features. Features from each residual block, except the first, are up-sampled via linear interpolation after a $1\times$1 convolution layer and combined with the features of the previous layer. Predicted indicators are generated by averaging outputs from a $3\times$3 convolutional layer and a fully connected layer. The clinical quantitative indicator discrimination module is continuously trained with input images throughout network training.

\subsubsection{Adversarial Discriminator}
The adversarial discriminator module designed to train the generator to distinguish between the generated image \(\widehat{y_j}\) and the real image \(y_j\),consists of five layers of convolutional neural networks. It combines the input image \(x_i\) with either \(\widehat{y_j}\) or \(y_j\) and processes the combined image to evaluate authenticity. The module outputs a \(28 \times 28\) matrix, with each value representing the authenticity of a corresponding pixel block.

\subsection{Loss Function}
The model incorporates adversarial loss, age-scaled pixel loss, and dynamic indicator loss into its composite loss function. Each loss is elaborated upon below.

\subsubsection{Adversarial Loss($\mathcal{L}_{adv}$)}
During adversarial training, the generator strives to produce $\widehat{y_{j}}$ that appears convincing enough to deceive the discriminator. The discriminator attempts to correctly distinguish between $\widehat{y_{j}}$ and $y_{j}$ T-GAN considers the ideas presented in PixelGAN \cite{isola2017image} to set up the loss functions. The computation of the loss of adversarial discrimination ${L}_{adv}^D$ is shown in Eq.(\ref{eq_loss_adv_D}), while the adversarial loss of the generator ${L}_{ad\nu}^G$  is shown in Eq.(\ref{eq_loss_adv_G}).

\begin{equation}
    \label{eq_loss_adv_D}
    \begin{aligned}
 \mathcal{L}_{adv}^D=-\log\left(D_{adv}\left(x_i\right)\right)\\-\log\left(1-D_{adv}\left(G(x_i,a_{j-i})\right)\right)
\end{aligned}
\end{equation}

\begin{equation}
    \label{eq_loss_adv_G}
    \mathcal{L}_{ad\nu}^G=-\log\left(D_{ad\nu}\left(G(x_i,a_{j-i})\right)\right)
\end{equation}
where $i, j$ represents the images of patient age periods $i, j$ respectively, and $G(x_i,a_{j-i})$ is $\widehat{y_{j}}$.

\subsubsection{Age-Scaled Pixel Loss($\mathcal{L}_{asp}$)}
To address the prevalent issue of an uneven distribution of time intervals in temporal brain image datasets, where the majority of the data consist of short-term sequences of three years or less, T-GAN employs a pixel loss with an age-scaling factor (referred to as $\mathcal{L}_{asp}$) to train the generator. The cosine distance between the input and target images is computed and then applied to the original loss $ell_1$ in order to increase the weight of the long-term (more than three years) image during training. The calculation formula for $\mathcal{L}_{asp}$ is shown in Eq.(\ref{eq_loss_age}):

\begin{equation}
    \label{eq_loss_age}    \mathcal{L}_{asp}=\mathbb{E}_{x_i,\widehat{y_j}}[\|x_i-\widehat{y_j}\|_1*\mathbb{C}(a_i,a_j)]
\end{equation}

where $\mathbb{C}(a_i,a_j)$ represents the age scaling factor, which is the cosine distance between the two vectors. The calculation of $\mathbb{C}(a_i,a_j)$ is done in Eq.(\ref{eq_age}):
\begin{equation}
    \label{eq_age}
    \mathbb{C}(a_i,a_j)=\cos\langle a_i,a_j\rangle=\frac{a_i^T a_j}{\|a_i\|_2\|a_j\|_2}
\end{equation}
where $a_i$ and $a_j$ represent the age encoding vectors for the input and target images, respectively.

\subsubsection{Dynamic Indicator Loss($\mathcal{L}_{dm}$)}
The indicator recorded in the dataset contains missing values due to several objective factors. To address this concern, T-GAN employs a dynamic indicator loss calculation function known as $\mathcal{L}_{dm}$. This function instructs the model by employing incomplete indicators and establishes an efficient quantitative indicator supervision mechanism. The calculation of the loss of the indicator discrimination $\mathcal{L}_{dm}^D$is shown in Eq.(\ref{eq_loss_di_D}), while the indicator loss of the generator $\mathcal{L}_{dm}^G$is shown in Eq.(\ref{eq_loss_di_G}).

\begin{equation}
    \label{eq_loss_di_D}    \mathcal{L}_{dm}^D=\sum_{p=1}^n\mathbb{I}(c_p\neq nan)\begin{Vmatrix}D_{index}\left(y_j\right)_p-c_p\end{Vmatrix}_1
\end{equation}

\begin{equation}
    \label{eq_loss_di_G}   \mathcal{L}_{dm}^G=\sum_{p=1}^n\mathbb{I}(c_p\neq nan)\left\|D_{index}\left(\hat{y}_j\right)_p-D_{index}\left(y_j\right)_p\right\|_1
\end{equation}

Where $c_p$ represents the $p$-th real clinical quantitative indicator, and $D_{index}\left(y_j\right)$ is the indicator predicted by the prediction module based on the input images. The symbol $\mathbb{I}(c_p\neq nan)$ denotes an indicator function that returns a value of 1 if a specific condition is true and 0 if it is false.

The total loss of the model is calculated by combining the loss functions from Eq. (\ref{eq_loss_adv_G}), Eq. (\ref{eq_loss_age}), and Eq. (\ref{eq_loss_di_G}). The computation for \(\mathcal{L}_{gan}\) is presented in Eq. (\ref{eq5}):

\begin{equation}
    \label{eq5}
\mathcal{L}_{gan}=\alpha\mathcal{L}_{adv}^G+\beta\mathcal{L}_{asp}+\gamma\mathcal{L}_{dm}^G
\end{equation}
where $\alpha$, $\beta$, and $\gamma$ are weight adjustment parameters for the respective losses, set to 1, 100, and 1.2.

\section{experimental setup}

This paper's experiments utilize data from the Alzheimer's Disease Neuroimaging Initiative (ADNI), a study that collects imaging data, biomarkers, and neuropsychological assessments related to Alzheimer's disease. This paper excluded samples with a single record to ensure that each sample has at least two records from distinct time points. The dataset includes 1,872 samples, primarily consisting of short temporal terms that span three years. Baseline checks consist of 1,872 records, with short-term follow-ups at 64.53\% (1,208 records) and extended periods at 35.47\% (664 records). Ages range from 55.0 to 97.3 years, with a median age of 75.3 years.

\subsection{Data Preprocessing}
\subsubsection{Image data preprocessing}
\begin{figure}[htbp]
    \centering
    \includegraphics[width=1\linewidth]{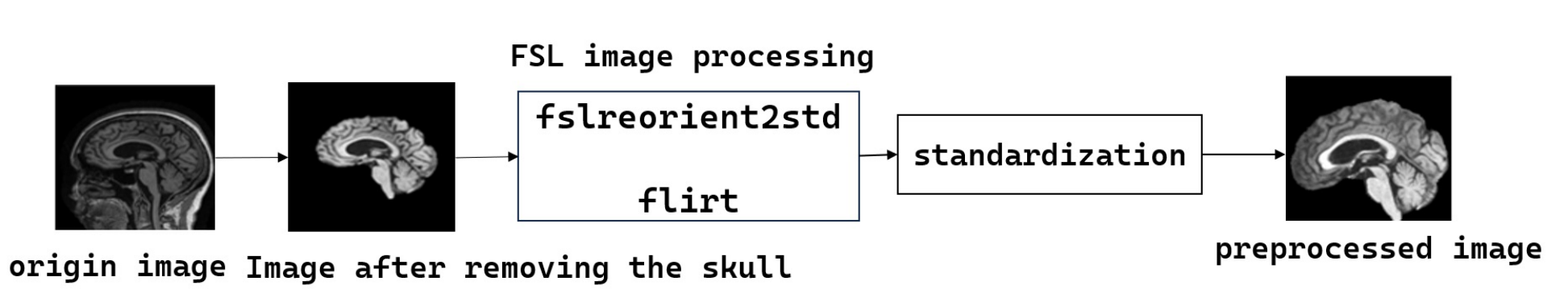}
    \caption{Image Preprocessing Procedure}
    \label{Image Preprocessing Procedure}
\end{figure}

The original MRI contains a lot of noisy information, such as the skull, nose, and other parts; it may also have a tilt and offset. This article uses a universal MRI pre-processing method to reduce noise and bias that may affect experimental results, as shown in Fig\ref{Image Preprocessing Procedure}. In order to prepare the image data, this paper used HD-BET\cite{isensee2019automated} to remove the skull and FSL\cite{woolrich2009bayesian} to correct head orientation. The images are then normalized and registered to the MNI152 template. The image size is scaled from 182x218x182 to 184x220x184 proportionally. The image is then resized to 220x220x220 by zero-filling. Lastly, the images are sliced in three distinct directions: axial, coronal, and sagittal.

\subsubsection{Preprocessing of clinical quantitative indicators}
\begin{figure}[htbp]
    \centering
    \includegraphics[width=1\linewidth]{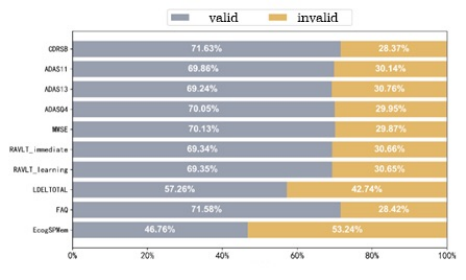}
    \caption{Proportion chart of missing clinical quantitative indicators selected}
    \label{proportion_indicators}
\end{figure}

The initial dataset comprised 36 indicators pertaining to AD. The paper utilized feature correlation analysis to determine the ten most relevant indicators for AD. The selected indicators had a decent amount of data completeness, with an average missing rate of 33.48\%, as depicted in Fig.\ref{proportion_indicators}.

\subsection{Evaluation Metrics}

In order to assess the quality of the generated image, various metrics were utilized, including Mean Absolute Error (MAE), Mean Squared Error (MSE), Structural Similarity Index (SSIM) \cite{wang2004image}, and Peak Signal-to-Noise Ratio (PSNR)

This study drew inspiration from previous work \cite{jung2020conditional, jung2021conditional} that used classification models to evaluate the quality of generated images. A novel scoring system, i.e., the Disease Feature Distance (DFD) score, was proposed to quantify the feature disparity between authentic and generated images. DFD is used to assess the degree of preservation of disease-related features in the generated images. The formula is expressed in the following manner in Eq.(\ref{eq:DFD}):

\begin{equation}
    \label{eq:DFD}
DFD=\sum_{i=1}^{N}\sqrt{(f(x_{gen}^{i})-f(x_{real}^{i}))^2}
\end{equation}

Wherein $N$ denotes the size of the test set, $i$ denotes the index of the $i$-th validation data, $x_{gen}$  and $x_{real}$ denote the generated and real images, respectively, and $f$ denotes the output feature following the fully connected layer of the classification model.

Quantitative indicators were evaluated using five metrics, i.e., Mean Absolute Error (MAE), Mean Squared Error (MSE), Mean Absolute Percentage Error (MAPE), Root Mean Square Error (RMSE), and Coefficient of Determination ($R^2$).

\section{experiment and discussion}
\vspace{1em}

The model was implemented with PyTorch 1.12 and Python 3.8 on a workstation with an Nvidia TITAN Xp GPU. The dataset was split into training and testing sets in a 9:1 ratio, and a five-fold cross-validation approach was used to calculate the mean and standard deviation of various metrics for comparison with other methods. All methods were trained using the provided dataset.

T-GAN training used   the Adam optimizer for 50 epochs, beginning with a learning rate of 1e-5. A cosine annealing strategy with a 30-epoch cycle and a minimum learning rate of 1e-8 was applied to adjust the learning rate during training.

\subsection{Ablation Experiment}
\subsubsection{Ablation of Quantitative Indicators}

The ADNI dataset contains 36 quantitative indicators related to Alzheimer's disease (AD), with a 52\% missing data rate.  This study used analysis of variance to rank the relationship between the indicators and AD labels, allowing for the selection of relevant indicators. As shown in Table \ref{tb:indicator}, two methods were employed: 'Max' for the sequential selection of the most relevant indicators and 'Random' for random selection.

Selecting relevant indicators improves model accuracy compared to random selection. The top 10 quantitative indicators identified in this study achieved the best results, with SSIM and PSNR values of 0.9158 and 26.38, respectively. The complexity of multi-indicator prediction increases with the number of indicators, suggesting that adding more may not be beneficial. Correlations between indicators and diseases vary, and highly correlated indicators can enhance image prediction. However, more indicators may lead to increased missing data, negatively affecting prediction accuracy and image quality.

\begin{table}[htbp]
\renewcommand\arraystretch{1.3}
   
    \caption{ablation of quantitative indicators}
    \centering
    \resizebox{\linewidth}{!}{
     \begin{threeparttable}

\begin{tabular}{c|c|c|c|c}
\hline
Selection Method                    & Number of indicators                      & MAE↓\tnote{1}                    & SSIM↑\tnote{2}                   & PSNR↑\tnote{2}               \\ \hline
\multirow{4}{*}{Select Best}        & Top 1                                     & \textbf{0.5578}                  & 0.8586                           & 25.83                           \\
                                    & Top 5                                     & 0.5876                           & 0.8625                           & 26.20                           \\
                                    & Top 10                                    & 0.6821                           & \textbf{0.9158}                  & \textbf{26.38} \\
                                    & Top 20                                    & 0.7215                           & 0.8584                           & 26.16                           \\ \hline
\multirow{2}{*}{Random}             & 5                                         & 0.6664                           & 0.8600                           & 25.75                           \\
                                    & 10                                        & 0.7281                           & 0.8582                           & 25.62                           \\ \hline
ALL                                 & 36                                        & 0.7200                           & 0.8611                           & 26.23                           \\ \hline
\end{tabular}

\begin{tablenotes}
       \footnotesize
       \item[1] The evaluation metric is used to assess the results of generating quantitative indicators.
       \item[2] The evaluation metric is used to assess the results of the prediction of MRI image.
     \end{tablenotes}
    \end{threeparttable}
}
\label{tb:indicator}
\end{table}

\subsubsection{Ablation of Loss Function}

This paper proposes two special loss functions for temporal image prediction, namely $\mathcal{L}_{asp}$ and $\mathcal{L}_{dm}$. The $\mathcal{L}_{asp}$ function improves the model's focus on long-term images, thereby addressing the issue of limited data volume for such images in the dataset. In contrast, a connection between quantitative indicators and image prediction is established by $\mathcal{L}_{dm}$, making it the most critical loss function.

For experiments in this section, the basic adversarial loss function $\mathcal{L}_{adv}$ was used by default. The outcomes of the ablation experiments for these loss functions are presented in Table \ref{tb:ablation}. The network demonstrated optimal performance when all loss functions were utilized, with $\mathcal{L}_{dm}$ delivering a more significant enhancement and enhancing accuracy by approximately 0.03 when compared individually using SSIM. In contrast, the value of $\mathcal{L}_{adv}$ was only marginally elevated by 0.01. The experimental findings indicate that utilizing both loss functions for all evaluation metrics, namely $\mathcal{L}_{asp}$ and $\mathcal{L}_{dm}$ , yields the optimal performance.

\begin{table*}[h]
\renewcommand\arraystretch{1.3}   
    \caption{Results of Ablation Experiments with Different Loss Combinations}
    \centering
    \resizebox{\linewidth}{!}{
\begin{tabular}{c|ccc|ccc|ccc}
\hline
\multicolumn{1}{c|}{\multirow{2}{*}{Loss}}                 
& \multicolumn{3}{c|}{Transverse Plane}                                                             
& \multicolumn{3}{c|}{Sagittal Plane}                                                               
& \multicolumn{3}{c}{Coronal Plane}                                                                 \\ \cline{2-10} 
\multicolumn{1}{c|}{}                                      
& \multicolumn{1}{c|}{MAE$\downarrow$}      & \multicolumn{1}{c|}{SSIM$\uparrow$}  & PSNR$\uparrow$
& \multicolumn{1}{c|}{MAE$\downarrow$}      & \multicolumn{1}{c|}{SSIM$\uparrow$}  & PSNR$\uparrow$ 
& \multicolumn{1}{c|}{MAE$\downarrow$}      & \multicolumn{1}{c|}{SSIM$\uparrow$}  & PSNR$\uparrow$ \\ \hline

$\mathcal{L}_{adv}$                                        
& \multicolumn{1}{c|}{0.04915}  & \multicolumn{1}{c|}{0.8493} & 25.17 
& \multicolumn{1}{c|}{0.04579}  & \multicolumn{1}{c|}{0.8217} & 24.04 
& \multicolumn{1}{c|}{0.04122} & \multicolumn{1}{c|}{0.8487} & 25.15 \\
$\mathcal{L}_{asp}$+ $\mathcal{L}_{adv}$                   
& \multicolumn{1}{c|}{0.03242}  & \multicolumn{1}{c|}{0.8715} & 23.67 
& \multicolumn{1}{c|}{0.02805}  & \multicolumn{1}{c|}{0.8661} & 24.72 
& \multicolumn{1}{c|}{0.02772}  & \multicolumn{1}{c|}{0.8881} & 25.91 \\

$\mathcal{L}_{dm}$+ $\mathcal{L}_{adv}$                    
& \multicolumn{1}{c|}{0.02549}  & \multicolumn{1}{c|}{0.8917} & 25.70 
& \multicolumn{1}{c|}{0.02394}  & \multicolumn{1}{c|}{0.9090} & 26.07 
& \multicolumn{1}{c|}{0.02345}  & \multicolumn{1}{c|}{0.9010} & 26.20 \\

$\mathcal{L}_{asp}$+$\mathcal{L}_{dm}$+$\mathcal{L}_{adv}$ 
& \multicolumn{1}{c|}{\textbf{0.02315}}  & \multicolumn{1}{c|}{\textbf{0.9115}} & \textbf{26.22} 
& \multicolumn{1}{c|}{\textbf{0.02235}} & \multicolumn{1}{c|}{\textbf{0.9058}} & \textbf{26.18} 
& \multicolumn{1}{c|}{\textbf{0.02356}}  & \multicolumn{1}{c|}{\textbf{0.9165}} & \textbf{26.24} \\ \hline
\end{tabular}
}
 \label{tb:ablation}
\end{table*}

\subsection{Experiment on Indicator Prediction}
The goal of T-GAN is to improve the quality of generated images by focusing on the accuracy of the predicted quantitative indicators generated by the discriminators indicator prediction branch. We evaluated this branch's performance by comparing it to several standard base networks \cite{he2016deep, howard2019searching, huang2017densely, SimonyanZ14a} and an age prediction network \cite{zhang2019c3ae}. In order to produce quantitative indicators, we modified the output layers of these networks to create a fully connected layer that gave a result of 10 indicators. The Dynamic Indicator Loss was used to train these models during the training process instead of traditional methods for filling in missing information. Our method demonstrated superior performance compared to the others, as shown in Table \ref{tb:ind} We had MAE, MSE, and R$^2$ values of 0.62008, 0.77543, and 0.33772, respectively. These results suggest that our method has a better fitting ability and smaller errors in quantitative indicator prediction.

\begin{table}[htbp]
\renewcommand\arraystretch{1.3}
    \caption{experiment on indicator prediction}
    \centering
\begin{tabular}{c|c|c}
\hline
Method & MAE$\downarrow$ & R$^2$$\uparrow$   \\ \hline
Resnet\cite{he2016deep}    &0.75360        &0.02167        \\ 
VGGnet\cite{2014Very} &0.68310        &0.14480        \\ 
C3AE\cite{zhang2019c3ae}    &0.75160      &0.12088        \\
Mobilnet\cite{howard2019searching}  &0.73619     &0.05938   \\ 
Densenet\cite{huang2017densely}  &0.70740 	 	  &0.13121   \\ 
T-GAN (our)  &\textbf{0.62008}   &\textbf{0.33772}  \\ \hline
\end{tabular}
\label{tb:ind}
\end{table}

\subsection{Experiment on MRI Image Prediction}
We select five GAN models for comparison in the temporal AD brain image prediction. The Age-ACGAN\cite{kan2020age},Pix2Pix-r\cite{isola2017image}, Pix2Pix-A\cite{isola2017image},The CAAE\cite{zhang2017age},The TR-GAN\cite{fan2022tr} .

\subsubsection{Evaluation of  Generated Images}

We conducted comparative experiments on image prediction, dividing the evaluation into short-term ( $\leq$ 3 years) and long-term(>3 years) groups. The evaluation metrics are presented in Table \ref{tb_img}. 

Our method achieved the highest MAE of 1.9471 in short-term image prediction. While TR-GAN and pix2pix-a had slightly better SSIM and PSNR scores, the differences were minimal at approximately 0.01 and 0.2, respectively. The Age-ACGAN and CAAE models scored lower in SSIM and PSNR due to the poor preservation of disease features.

Long-term image prediction results showed that pix2pix-a had the highest PSNR, exceeding our method by approximately 0.04. Nevertheless, our methodology outperformed others in several metrics, with an SSIM superior to Pix2Pix-r by approximately 0.02.Models with higher image prediction quality extracted features that were more closely aligned with real images. In the DFD (AD/MCI) experiments, our method slightly lagged behind pix2pix-a for short-term predictions, but the difference was negligible.

The experimental results show that most methods yield higher quality short-term predictions than long-term ones. The ADNI dataset has more short-term data, which may limit the availability of long-term data. Nonetheless, our method effectively generates long-term images using the age-scaling mechanism $\mathcal{L}_{asp}$ and quantitative indicators $\mathcal{L}_{dm}$, indicating an accurate reflection of the characteristics of disease progression.

\begin{table*}[htbp]
\renewcommand\arraystretch{1.0}    
    \caption{Comparative Experiment on Image Prediction}
    \centering

 \begin{tabular}{c|c|c|c|c|c|c}
\hline
Term & Method & MAE $\times10^{-2}$ ↓ & SSIM ↑  & PSNR ↑ & \makecell{DFD(AD/CN)}↓ & \makecell{DFD(AD/MCI)}↓\\
\hline
\multirow{6}{*}{\makecell{Short\\Term\\($\leq$ 3 years)}}
& Age-ACGAN\cite{kan2020age} & 8.6670$_{\pm1.852}$ &0.622$_{\pm0.050}$  & 15.137$_{\pm1.367}$ & 0.4369   & 0.3602                                                                                                       \\
& Pix2Pix-a\cite{isola2017image}& 4.1036$_{\pm0.088}$  & 0.895$_{\pm0.001}$ &\textbf{27.806$_{\pm\textbf{0.252}}$}& 0.3824 & \textbf{0.3447}                                                                                                                  \\
        & Pix2Pix-r\cite{isola2017image}
         & 2.5179$_{\pm0.078}$                                     & 0.898$_{\pm0.004}$                   & 25.501$_{\pm0.378}$
         & 0.3857                                & 0.3922                                                                                                                            \\
         & CAAE\cite{zhang2017age}
         & 11.6085$_{\pm0.101}$                                    & 0.673$_{\pm0.002}$                   & 18.386$_{\pm0.087}$
         & 0.4954                                & 0.5291                                                                                                                      \\
         & TR-GAN\cite{fan2022tr}
         & 2.4277$_{\pm0.191}$                                     & \textbf{0.943$_{\pm\textbf{0.010}}$} & 26.874$_{\pm0.554}$
         & 0.4158                                & 0.3771                                                                                                                     \\
         & T-GAN(our)
         & \textbf{1.9471$_{\pm\textbf{0.016}}$}  & 0.935$_{\pm0.002}$                   & 27.593$_{\pm0.077}$
         & \textbf{0.3805}                       & 0.3453                                                                                                                    \\
        \hline
        \multirow{6}{*}{\makecell{Long\\Term\\($>$3 years)}}
         & Age-ACGAN\cite{kan2020age} 
         & 8.7123$_{\pm1.816}$  & 0.621$_{\pm0.049}$    & 15.122$_{\pm1.342}$
         & 0.4106    & 0.3887                                                                                                                     \\
         & Pix2Pix-a\cite{isola2017image}
         & 6.7565$_{\pm0.150}$                                     & 0.824$_{\pm0.004}$                   & \textbf{23.721$_{\pm\textbf{0.157}}$}
         &  0.3637                               & 0.3725                                                                                                                      \\
         & Pix2Pix-r\cite{isola2017image}
         & 3.4408$_{\pm0.074}$                                    & 0.864$_{\pm0.004}$                   & 22.948$_{\pm0.221}$
         &  0.3777                               & 0.3901                       
         \\
         & CAAE\cite{zhang2017age}
         & 11.4915$_{\pm0.091}$                                     &0.674$_{\pm0.001}$ & 18.491$_{\pm0.068}$
         &  0.4997 & 0.4830 
         
\\
& TR-GAN\cite{fan2022tr}& 6.7226$_{\pm0.380}$  &0.817$_{\pm0.018}$ & 17.729$_{\pm0.529}$ & 0.4116  & 0.3878   

\\
& T-GAN(our) & \textbf{3.1294$_{\pm\textbf{0.074}}$} &\textbf{0.882$_{\pm\textbf{0.003}}$} & 23.683$_{\pm0.149}$& \textbf{0.3608} & \textbf{0.3681}                                                                                                            \\
        \hline
    \end{tabular}
    
    \label{tb_img}
\end{table*}

\subsubsection{Experiment on Sequential Prediction of MRI Images}

\begin{figure}[htbp]
    \centering
   
    \subfloat[Generated images ($|\widehat{y_j}|$)]{
        \includegraphics[width=1\linewidth]{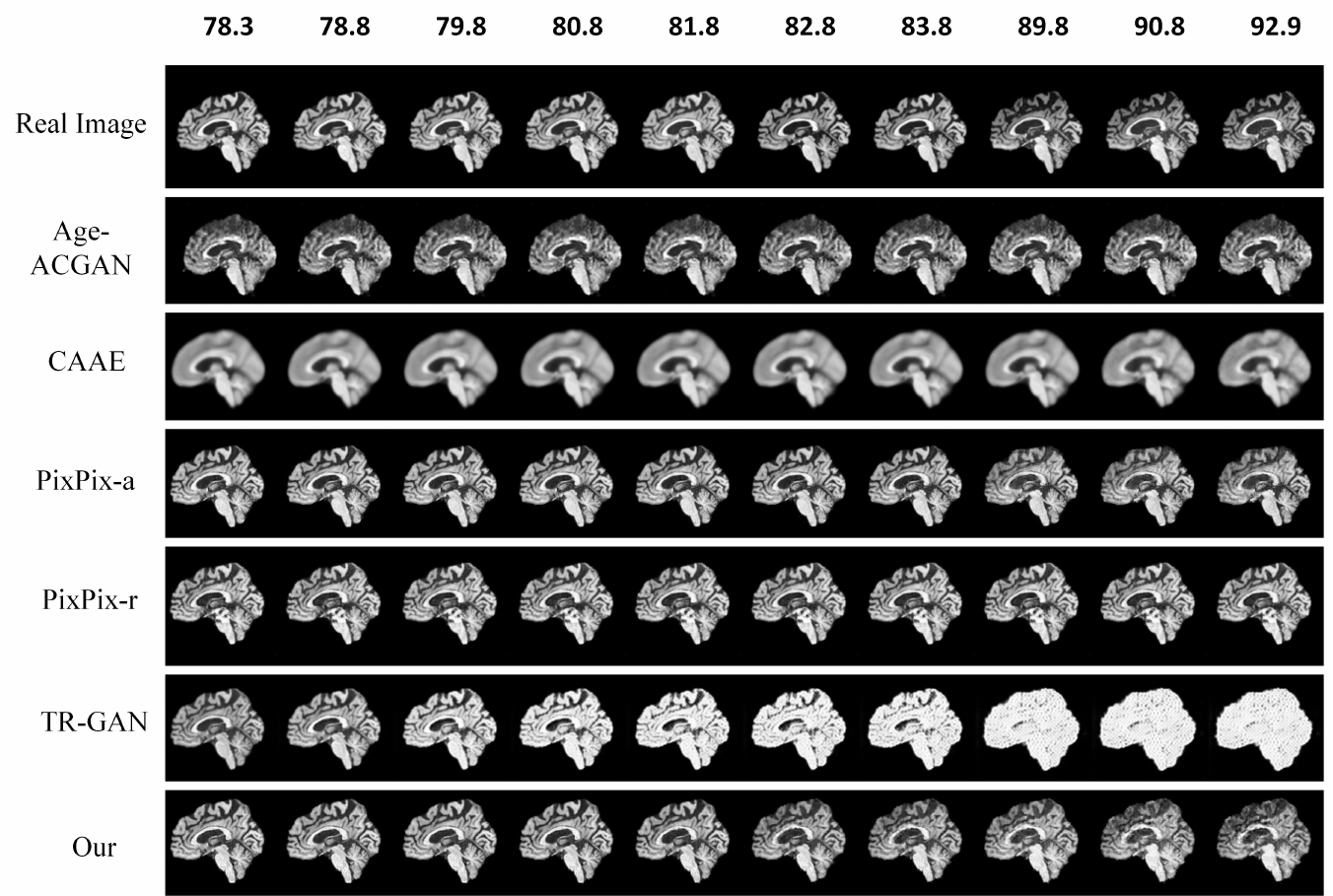}
    }\hspace{-0.12in}
    \subfloat[Temporal differences ($|\widehat{y_j}-x_i|$)]{
        \includegraphics[width=1\linewidth]{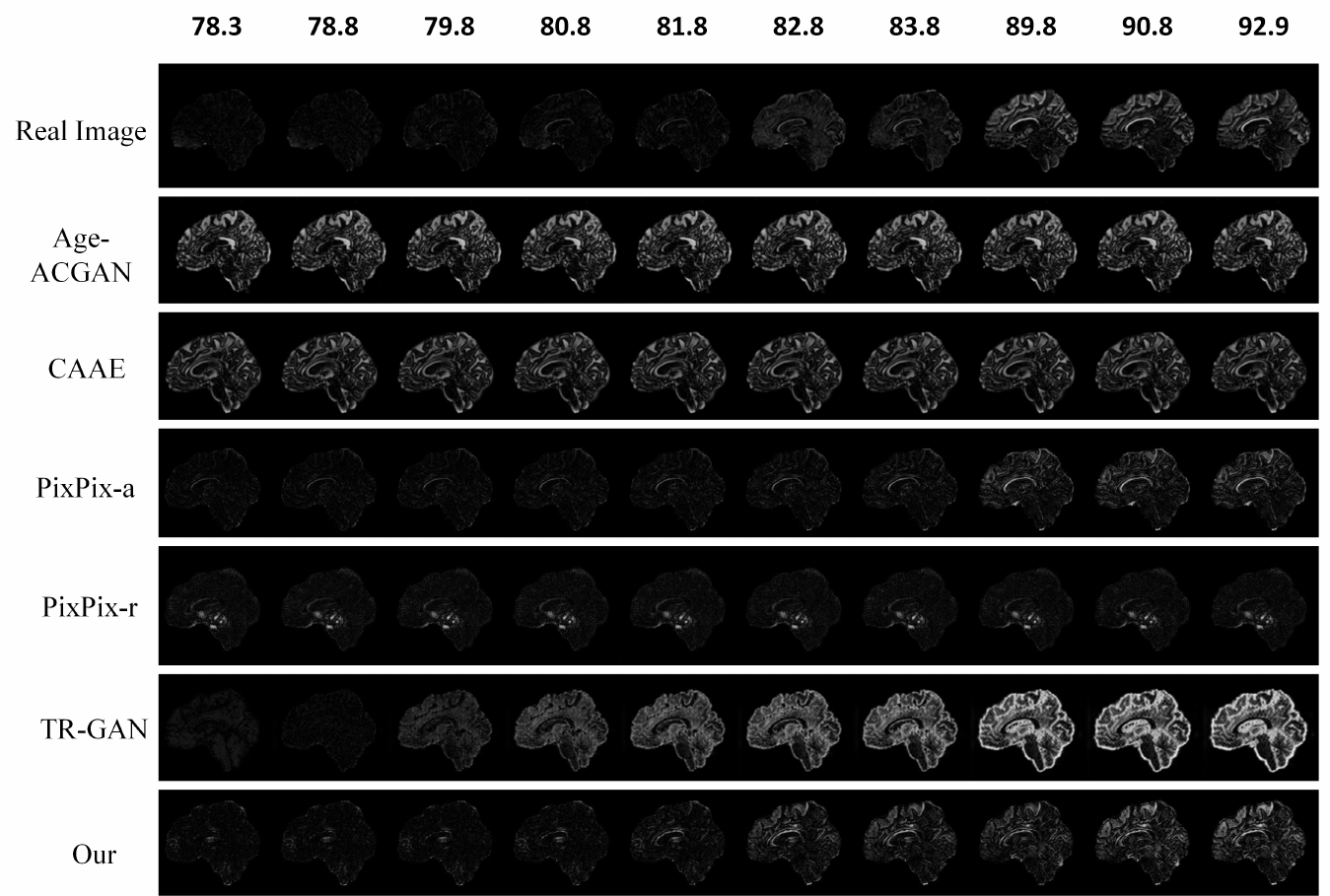}
    }\hspace{-0.12in}
    \subfloat[Horizontal differences ($|\widehat{y_j}-y_j|$)]{
        \includegraphics[width=1\linewidth]{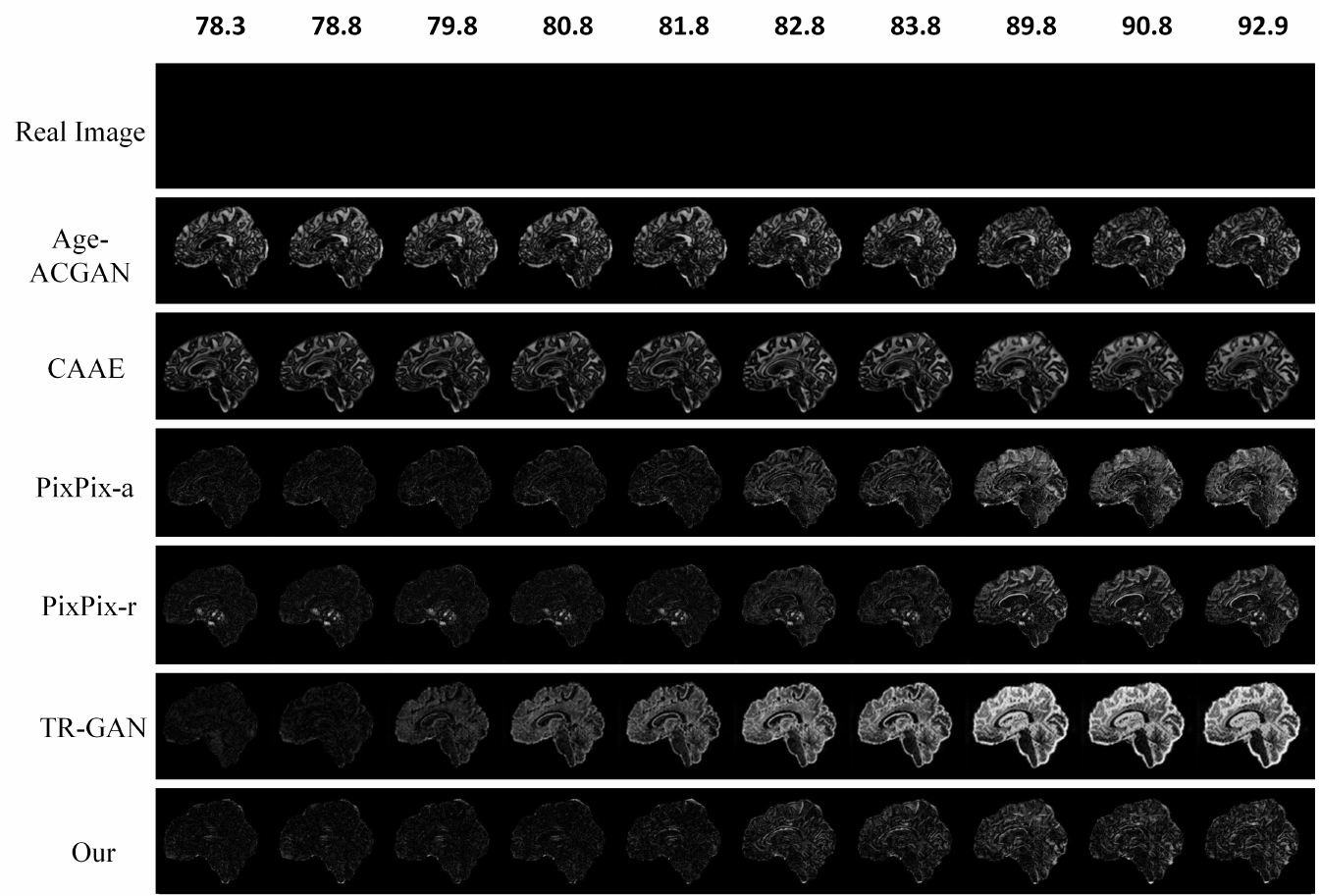}
    }\hspace{-0.12in}
   
    \caption{MRI image sequence prediction results in temporal and horizontal comparison with difference maps. The sample includes all images of patient `126\_S\_0680' (aged between 77.8 and 92.9 years). Among them, the image at 77.8 years old is used as the input for all models. (a) shows the MRI image sequences generated by each model ($|\widehat{y_j}|$); (b) shows the sequence differences in temporal MRI images ($|\widehat{y_j}-x_i|$); (c) shows the transversal difference in the same period in the sequence ($|\widehat{y_j}-y_j|$).}
    \label{fig_sr}
\end{figure}

Fig \ref{fig_sr} compares six models for generating sequences of brain images: (a) \( |\widehat{y_j}| \): generated sequential MRI images; (b) \( |\widehat{y_j}-x_i| \): differences from the initial MRI images; (c) \( |\widehat{y_j}-y_j| \): differences from the actual temporal MRI images.

The experimental results suggest that Age-ACGAN, CAAE, and Pix2Pix-r produce images of different ages with minimal differences. However, they failed to accurately reflect the features of disease progression. Pix2pix-r produced decent prediction results, but it lacked accurate age information, making it difficult to generate brain images corresponding to the intended periods. Despite the improvements made by Pix2pix-a, the learned sequence change features remain limited. TR-GAN has also observed this phenomenon, exhibiting lower-quality outcomes for generating long-term results. Our method relies on an attention-based conditional fusion mechanism, which enables it to detect detailed sequence changes in temporal images. It results in image sequences that closely resemble real ones and adapt better to the changing trends of real images.

\subsubsection{Experiment on  Age-Specific Generated MRI Images}
\begin{figure*}[htbp]
    \centerline{\includegraphics[width=0.8\linewidth]{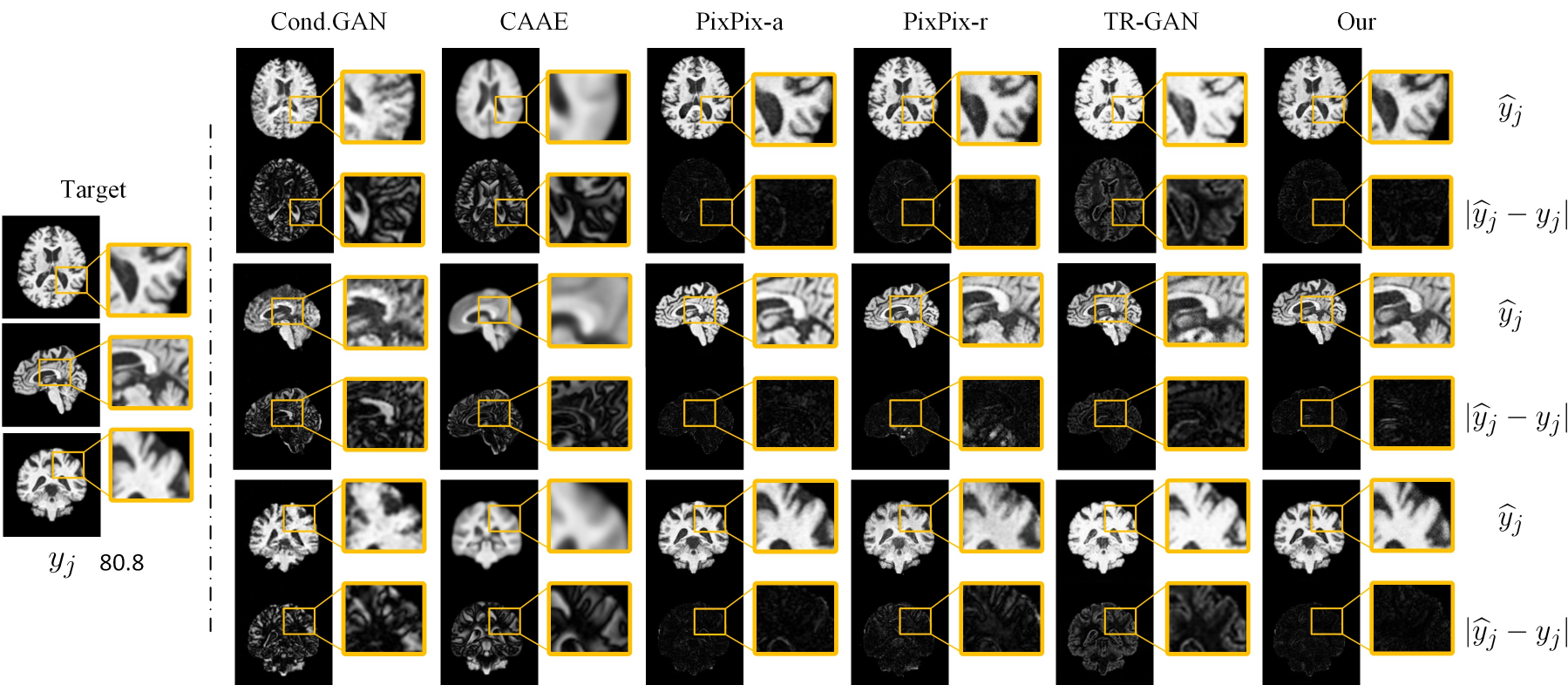}}
    \caption{ The data is from patient ``126\_S\_0680", with the image at 77.8 years as an input and the images at 80.8 years as the targets.}
    \label{fig_cr} 
\end{figure*}

 The particulars depicted in the Age-Specific generated images are compared through a cross-sectional perspective in Fig. \ref{fig_cr}. Each group of images from diverse models comprises slices from three distinct directions: namely, the axial, coronal, and sagittal planes. The upper row of each group displays the generated images, whereas the second row displays the distinction maps between the generated and genuine images. Furthermore, the first column represents the overall image, and the second column provides a zoomed-in detail of the image.

Both the Age-ACGAN and CAAE models encountered difficulties in generating high-quality and detailed MRI images due to deficiencies in learning intricate features. The output from the TR-GAN displayed specific deviations in detail, which can be attributed to the greater number of iterations involved in the process. On the contrary, the Pix2Pix-a, Pix2Pix-r, and T-GAN devices produced images of exceptional quality. Nonetheless, our approach stands out due to its exceptional degree of precision. We have introduced quantitative indicators for supervision, commonly referred to as $\mathcal{L}_{dm}$, which enable T-GAN to perceive pathology details associated with these indicators. This particular focus on disease characteristics significantly enhances the quality of the final details generated by the models.

\subsubsection{Experiment on Generated PET Images}
\begin{figure}[htbp]
    \centerline{\includegraphics[width=0.9\linewidth]{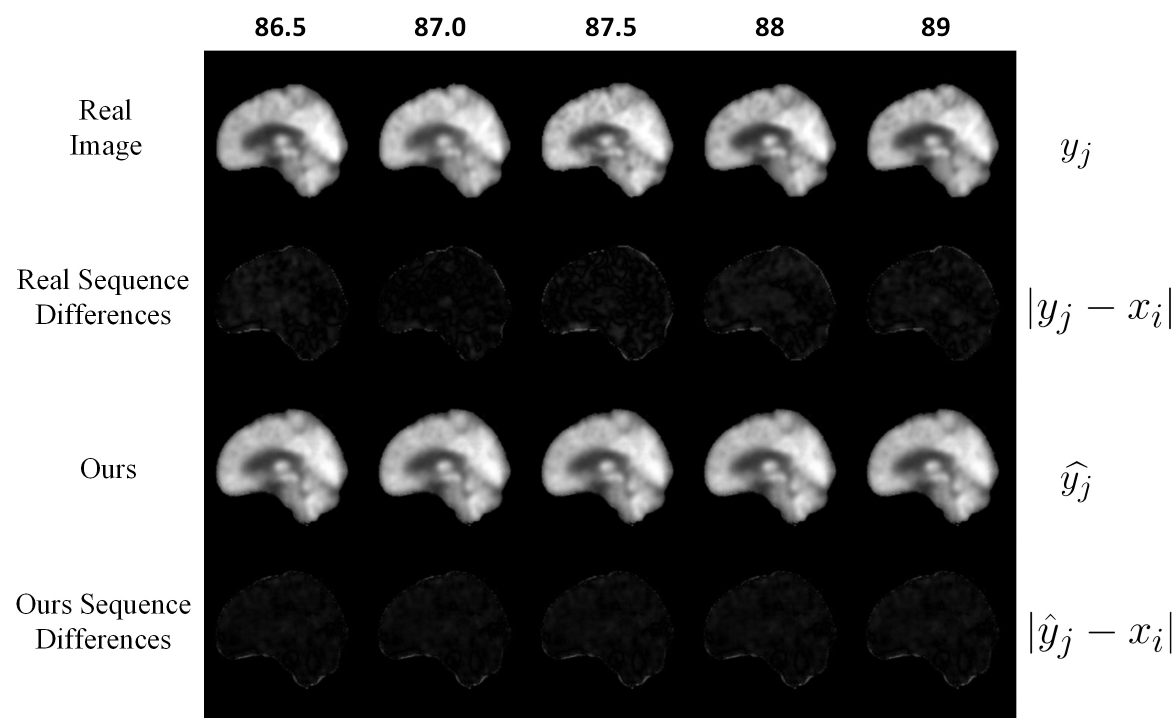}}
    \caption{PET image sequence generation results. The data is from patient ``003\_S\_1074", with the image at 86 years as an input and the images at 86.5 and 89 years as the targets.}
    \label{fig_pet}
\end{figure}

In order to further validate the generalization capability of the method, this chapter refines the model with PET images and then verifies its generative capability on PET. As depicted in Fig. \ref{fig_pet}, the contrast between the PET-generated sequence and the genuine sequence is illustrated.

It is evident that the generated image  by T-GAN closely resembles the genuine image in terms of morphology and structure. Furthermore, by comparing the longitudinal variance between the two sequences, it can be observed that the generated sequence exhibits similarity in longitudinal variance to the genuine image. The experimental results confirm that T-GAN  is effective in generating brain images in different modalities. The SSIM of the generated image with the real image in PET image generation experiments reaches 0.915, and the PSNR reaches 30.33.

\section{Conclusion}
This paper introduces T-GAN, a novel temporal brain image prediction adversarial network. T-GAN is capable of generating both quantitative indicators and image data simultaneously, thereby facilitating the prediction of future brain images and corresponding quantitative indicators for AD patients over a specific period. T-GAN reduces the issue of an imbalanced time interval distribution in the dataset compared to other models. The age-scaled pixel loss $\mathcal{L}_{asp}$ and the attention-based age-conditional fusion mechanism aim to achieve optimal long-term MRI image prediction outcomes. Furthermore, in comparison experiments for generating details, T-GAN benefits from the additional quantitative indicator prediction discriminant branch and dynamic indicator loss  $\mathcal{L}_{dm}$, exhibiting a more powerful detail prediction capability. These mechanisms establish a link between quantitative indicators and image prediction, effectively guiding the generator to learn the pathological information embedded in the quantitative indicators. Overall, T-GAN combines MRI images with quantitative indicator prediction, preserving disease features in the generated MRI images.

\section*{Acknowledgments}
This work is supported in part by the Natural Science Foundation of Fujian Province, China under Grant 2022J01318; in part by the Scientific Research Start-up Fund Project for High-level Researchers of Huaqiao University under Grant 22BS105.

\bibliography{refs0108}{}

\end{document}